\documentclass[10pt,twocolumn,letterpaper]{article}

\usepackage[pagenumbers]{cvpr} 

\usepackage{graphicx}
\usepackage{amsmath}
\usepackage{amssymb}
\usepackage{booktabs}

\usepackage{bm}
\usepackage[pagebackref,breaklinks,colorlinks]{hyperref}

\usepackage[capitalize]{cleveref}
\crefname{section}{Sec.}{Secs.}
\Crefname{section}{Section}{Sections}
\Crefname{table}{Table}{Tables}
\crefname{table}{Tab.}{Tabs.}

\usepackage{xcolor}

\newcommand{\ieno}{\textit{i}.\textit{e}.}
\newcommand{\egno}{\textit{e}.\textit{g}.} 

\def\cvprPaperID{4122} 
\def\confName{CVPR}
\def\confYear{2022}

\newlength\savewidth\newcommand\shline{\noalign{\global\savewidth\arrayrulewidth
  \global\arrayrulewidth 1pt}\hline\noalign{\global\arrayrulewidth\savewidth}}

\usepackage{array}
\newcolumntype{C}[1]{>{\centering\arraybackslash}m{#1}}
\newcolumntype{R}[1]{>{\raggedleft\arraybackslash}m{#1}}
\newcolumntype{P}[1]{>{\raggedright\arraybackslash}p{#1}}
\newcolumntype{M}[1]{>{\centering\arraybackslash}m{#1}}
\usepackage{multirow}

\begin{document}

\title{Unleashing the Potential of Unsupervised Pre-Training with Intra-Identity Regularization for Person Re-Identification}

\author{Zizheng Yang \qquad Xin Jin \qquad Kecheng Zheng \qquad Feng Zhao\thanks{Corresponding Author.}\\
University of Science and Technology of China\\
{\tt\small yzz6000@mail.ustc.edu.cn}
}

\maketitle

\begin{abstract}




Existing person re-identification (ReID) methods typically directly load the pre-trained ImageNet weights for initialization. However, as a fine-grained classification task, ReID is more challenging and exists a large domain gap between ImageNet classification. Inspired by the great success of self-supervised representation learning with contrastive objectives, in this paper, we design an Unsupervised Pre-training framework for ReID based on the contrastive learning (CL) pipeline, dubbed UP-ReID. During the pre-training, we attempt to address two critical issues for learning fine-grained ReID features: (1) the augmentations in CL pipeline may distort the discriminative clues in person images. (2) the fine-grained local features of person images are not fully-explored. Therefore, we introduce an \textbf{intra-identity} (I$^2$-)regularization in the UP-ReID, which is instantiated as two constraints coming from global image aspect and local patch aspect: a global consistency is enforced between augmented and original person images to increase robustness to augmentation, while an intrinsic contrastive constraint among local patches of each image is employed to fully explore the local discriminative clues. Extensive experiments on multiple popular Re-ID datasets, including PersonX, Market1501, CUHK03, and MSMT17, demonstrate that our UP-ReID pre-trained model can significantly benefit the downstream ReID fine-tuning and achieve state-of-the-art performance. Codes and models will be released to \url{https://github.com/Frost-Yang-99/UP-ReID}.




\end{abstract}

\section{Introduction}
\label{sec:intro}

\begin{figure}[ht]
	\centering
	\begin{subfigure}{.45\textwidth}
		\centering
		\includegraphics[width=\textwidth]{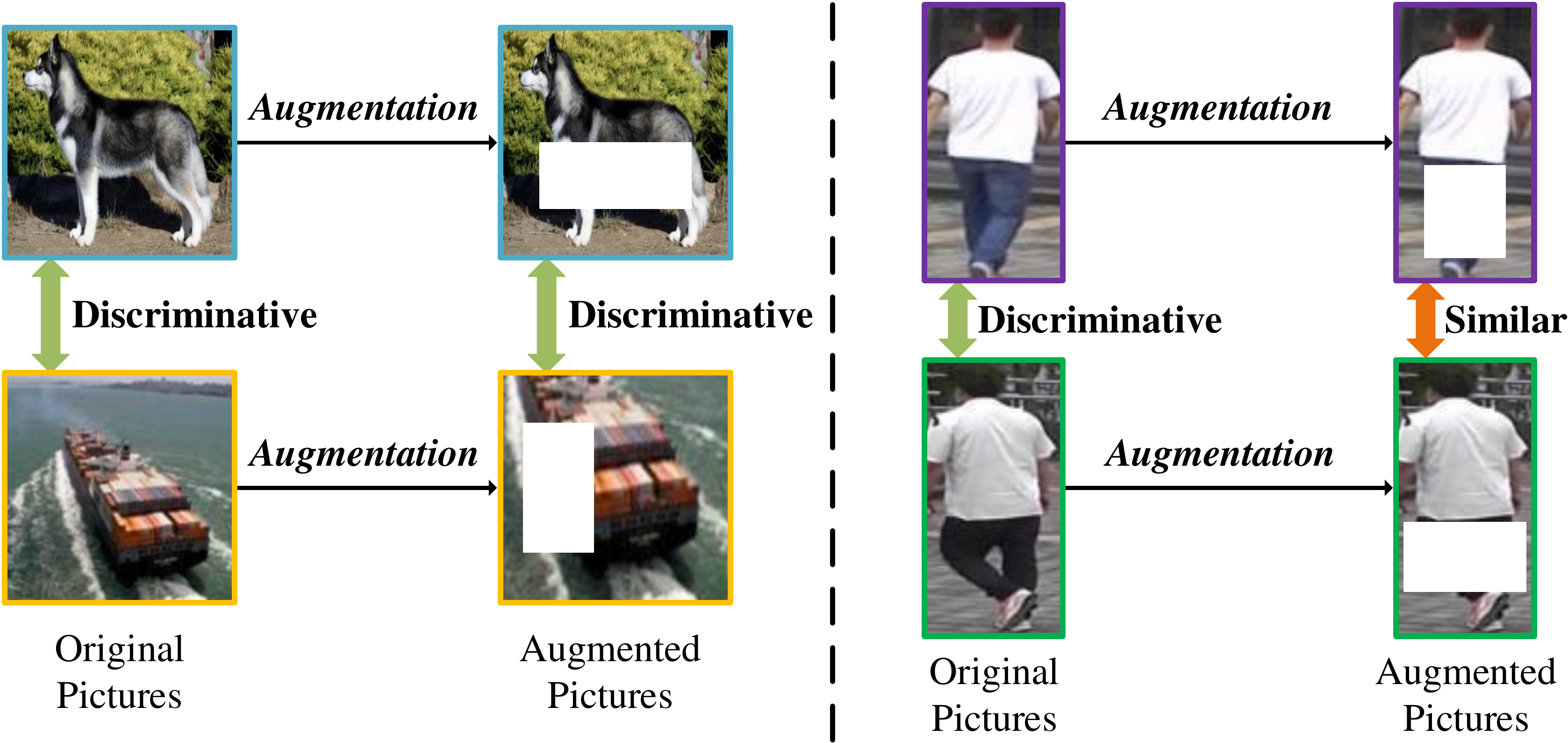}
		\caption{Left: the two augmented images are still discriminative for general classification tasks. Right: the discriminative attributes of two person images are ruined by augmentation for person ReID.}
		\label{fig:augmentation}
	\end{subfigure}%
	\vspace{2mm}
	\begin{subfigure}{.475\textwidth}
		\centering
		\includegraphics[width=\textwidth]{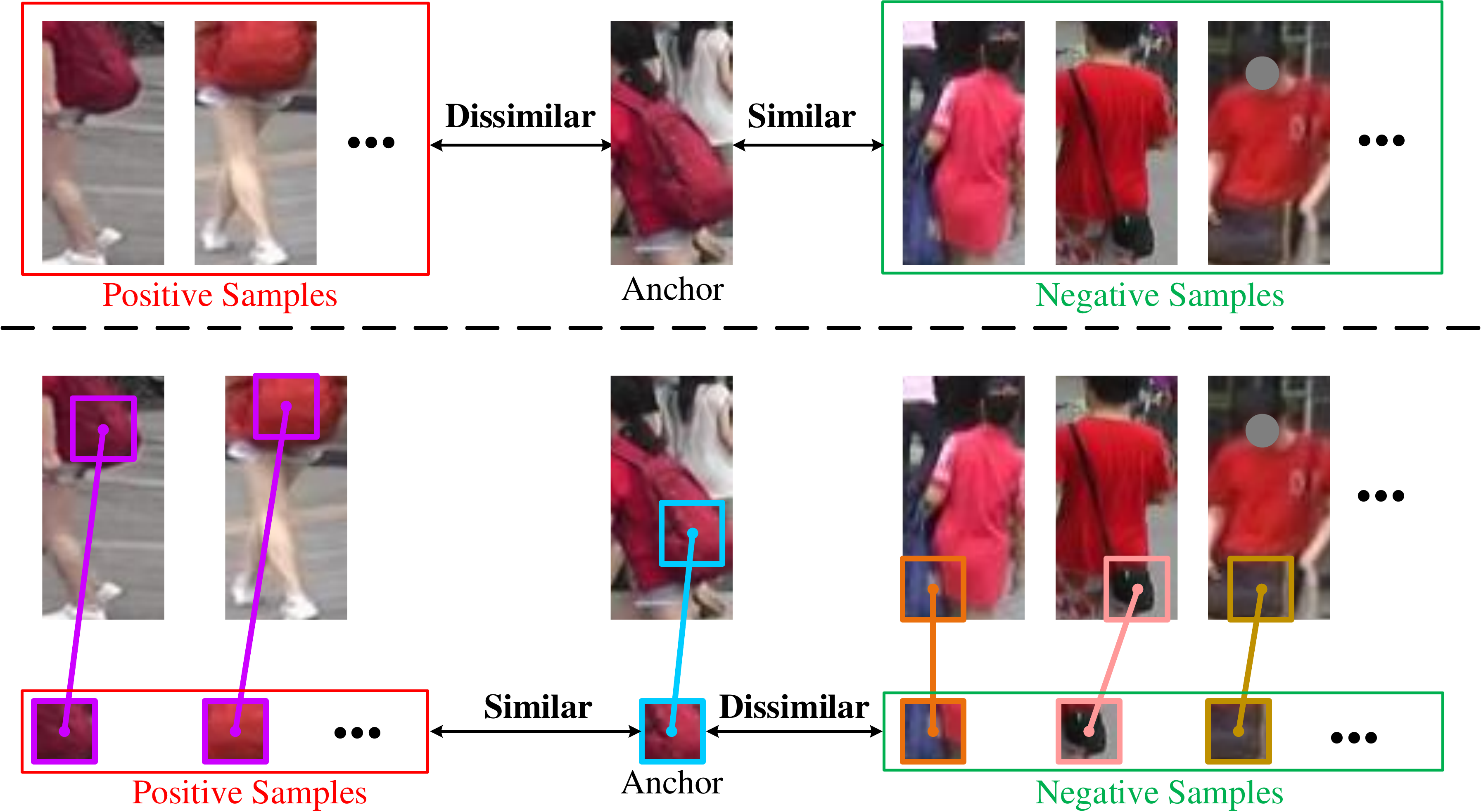}
		\caption{Top: a case that uses the global features for a failed ReID, where the positive samples share dissimilar appearance but the negative samples have a similar appearance instead. Bottom: a case that uses the fine-grained discriminative attributes, such as backpacks and bags, for a successful ReID where the person images are distinguishable and independent to clothing.}
		\label{fig:fine-grained}
	\end{subfigure}
	\vspace{-3mm}
	\caption{Two critical issues in the existing contrastive learning based pre-training methods, which should be well solved in the ReID-specific pre-training framework.}
	\vspace{-3mm}
\end{figure}

As a fine-grained classification problem, person re-identification (ReID) aims at identifying a specific person across non-overlapping camera views. Existing ReID methods have achieved a remarkable success in both supervised ~\cite{wang2018learning,suh2018part,shen2018person,zheng2021pose,zhang2019densely,jin2020uncertainty,jin2020semantics} and unsupervised~\cite{yang2019patch,lin2019bottom,ge2020mutual,ge2020self,jin2020style,dai2021cluster} domains. Most of these approaches directly leverage the weights pre-trained on ImageNet for model initialization, which is not optimal for ReID tasks, resulting in poor fine-tuning performance and slow convergence~\cite{wang2018learning,ge2020self}. The main reasons stem from two aspects: inapplicable pre-training method (ImageNet is more like a coarse-grained classification), and large domain gap between ImageNet and ReID datasets. Thus, how to efficiently pre-train a good ReID-specific initialization network is still under-explored.



Unsupervised pre-training has achieved a fast development with the great success of contrastive learning~\cite{he2020momentum,chen2020improved,chen2020simple,chen2021exploring,caron2020unsupervised}, which is taken as a pretext work, serving for different downstream supervised or unsupervised ReID fine-tuning algorithms. Going beyond the general pre-training task, this paper aims to propose a ReID-specific pre-training framework (\eg, pre-training a ResNet50~\cite{he2016deep} for learning discriminative ReID representations) on a large-scale unlabeled dataset. The pioneering work of~\cite{fu2021unsupervised} makes the first attempt on ReID pre-training and introduces a new large-scale unlabeled ReID dataset LUPerson. However, it directly transfers the general pre-training process based on contrastive learning that designed for ImageNet classification to ReID task, which ignores the fact that ReID is a fine-grained classification problem. This solution faces the following two critical issues:


The first issue comes from the augmentations used in the existing contrastive learning pipeline, which could possibly damage the discriminative attributes of person images. As shown in Figure~\ref{fig:augmentation}, different from the coarse-grained classification problem on ImageNet, the discriminative attributes of person images are prone to be destroyed by the augmentation operations. For example, in the ImageNet classification, although the augmentations applied to the pictures~(\eg, dogs and ships) may cause the lack of regional information, the remaining parts are still discriminative enough to support the model for distinguishing them. 
However, when applying the same augmentations to person images in ReID, it will cause a disaster---the most discriminative attributes (\ieno, trousers color) of person images are destructive, making them indistinguishable.



The second issue is that the fine-grained information of person images is not fully explored in previous pre-training methods. They typically only care about the learning of image-level global feature representations. Nevertheless, as a fine-grained classification task, ReID needs detailed local features in addition to global ones for the accurate identity matching~\cite{wang2018learning,yang2019patch,sun2018beyond}. As illustrated in Figure~\ref{fig:fine-grained}, the local fine-grained clues (\egno, backpacks, cross-body bags) are more helpful than global features w.r.t distinguishing different persons.


To address the above issues, we introduce an \emph{intra-identity (I$^2$-)} regularization in our proposed ReID-specific pre-training framework UP-ReID. It consists of a \emph{global consistency} constraint between augmented and original person images, and an \emph{intrinsic contrastive} constraint among local patches of each image.
Specifically, we first enforce a global {consistency} to make the pre-training model be more invariant to augmentations. We feed the augmented images as well as the original images into the model and then narrow the similarity distance between them in distributions.
Second, we propose an \emph{intrinsic contrastive} constraint for the local information exploration. Instead of directly feeding the holistic augmented images, we partition them into multiple patches and then send these patches along with the holistic images to the network. After that, We compute an intrinsic contrastive loss among patches to encourage the model to learn both fine-grained and semantic-aware representations. Moreover, based on the prior knowledge that human body is horizontally symmetric, we establish a hard mining strategy for the calculation of this loss, which makes the training stable and thus improves the generalization ability of the pre-trained model.


We summarize our main contributions as follows:

\begin{itemize}

\item To the best of our knowledge, this is the first attempt toward a ReID-specific pre-training framework dubbed UP-ReID by explicitly pinpointing the difference between the general pre-training and ReID pre-training.


\item Considering the particularity of ReID task, we introduce an intra-identity (I$^2$-)regularization in our ReID pre-training framework UP-ReID, which is instantiated from the global image level and local patch level.


\item In the I$^2$-regularization, a global consistency is first enforced to increase the robustness of pre-training to data augmentations. An intrinsic contrastive constraint with prior-based hard mining strategy among local patches of person images is further introduced to fully explore the local discriminative clues.



\end{itemize}

Extensive experiments on multiple widely-used ReID benchmarks demonstrate the effectiveness of the proposed pre-training framework UP-ReID. It outperforms the state-of-the-art pre-training methods by prominent margins, and could benefit a series of downstream ReID-related tasks.

\begin{figure*}[t!]
\centering
\centering
\includegraphics[width=16.5cm, height=10.5cm]{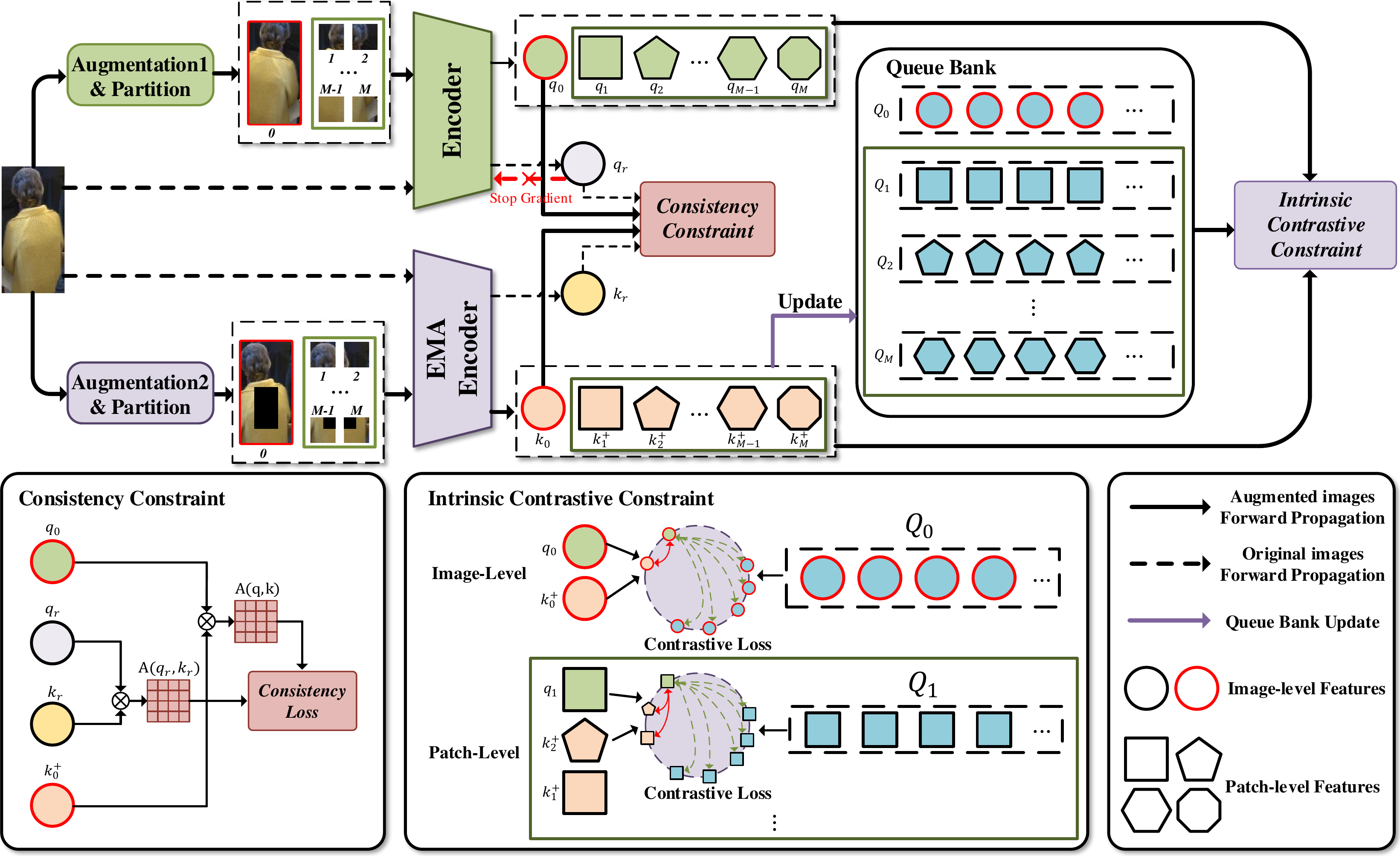}
\caption{Architecture of the proposed UP-ReID. Given an input image, we can get two different groups of augmented instances after two different augmentations and partition. Then, we feed them into the online encoder and EMA encoder respectively, together with the original images. A consistency loss is computed to narrow the gap between the similarity distribution of the augmented images and that of the original images. We also compute an intrinsic contrastive loss based on a delicately designed hard mining strategy. The EMA encoder features are used to update the queue bank. The online encoder is optimized by the gradient of the total loss, while the EMA encoder is updated by momentum-based moving average of the online encoder.}
\label{fig:Framework}
\vspace{-2mm}
\end{figure*}

\section{Related Work}

\subsection{Person ReID}

\noindent\textbf{Fully-supervised ReID approaches.} Fully-supervised ReID methods are based on supervised learning with labeled datasets and have achieved a great success~\cite{li2014deepreid,shen2018person,zhang2019densely,jin2020semantics}. These works can be divided into two mainstream branches. One focuses on designing effective optimization metrics (i.e., metric learning) for person ReID, such as hard triplet loss~\cite{hermans2017defense} and circle loss~\cite{sun2020circle}. On the other hand, learning fine-grained features is also a popular branch. PCB~\cite{sun2018beyond} and MGN~\cite{wang2018learning} both leverage local features of pedestrian images by manually splitting each holistic image into multiple sub-parts to achieve accurate person ReID. These methods are limited by the large-scale annotations and cannot be directly applied to unlabeled datasets.

\noindent\textbf{Unsupervised ReID approaches.} There are two typical categories of unsupervised person ReID: Unsupervised Domain Adaptation (UDA) based methods and Domain Generalization (DG) based methods. 1) UDA could handle the domain gap issue when the target domain data are accessible, which aims to learn a generic model from both labeled source data and unlabeled target data. The UDA-based methods can be further categorized into three main classes: style transfer based works~\cite{deng2018image,wei2018person,zheng2019joint}, attribute recognition based works~\cite{wang2018transferable,qi2019novel} and pseudo labeling based works~\cite{song2020unsupervised,ge2020mutual,ge2020self}. 2) DG is designed for a more challenging case where the target domain data are unavailable. Jin etal~\cite{jin2020style} designs a Style Normalization and Restitution(SNR) module to enhance the identity-relevant features and filter out the identity-irrelevant features for improving the model's generalization ability. In addition, meta-learning~\cite{zhao2021learning} is also employed as a popular way to achieve person ReID specific domain generalization. However, all these methods generally load the pre-trained ImageNet weights for initialization, and ignore the gap between the ImageNet classification and the fine-grained ReID task.


\subsection{Self-Supervised Representation Learning}

Based on the recently popular contrastive learning, the unsupervised pre-training has achieved a great success, and many representative works have achieved comparable or even slightly better performance than supervised works. Moco~\cite{he2020momentum} and Moco v2~\cite{chen2020improved} design a dynamic queue and introduce a momentum update mechanism to optimize a key encoder progressively. SimCLR\cite{chen2020simple} and SimCLR v2~\cite{chen2020big} also achieve great performance with a large batch size, rich data augmentations and a simple but effective projection head. BYOL~\cite{grill2020bootstrap} and SimSiam~\cite{chen2021exploring} further achieve great performance even without negative pairs. SwAV~\cite{caron2020unsupervised} replaces comparison between pairwise samples with comparison between cluster assignments of multiple views.

The work of~\cite{fu2021unsupervised} proposes a new large-scale unlabeled dataset ``LUPerson'' which is large enough to support pre-training and makes the first attempt to pre-train specific models for person ReID initialization. However, since the work merely migrates the approach of pre-training models on ImageNet to ReID directly, it suffers from the instability issue (see Figure~\ref{fig:augmentation}) caused by augmentation and lacked of the exploration of fine-grained discriminative information of pedestrian images (see Figure~\ref{fig:fine-grained}). 


\section{Unsupervised Pre-training for ReID}

Person ReID training typically contains two procedures of pre-training and fine-tuning: (a) the model~(\eg, ResNet50) is first \emph{pre-trained} unsupervisedly on a large-scale dataset~(\eg, LUPerson~\cite{fu2021unsupervised}) with a pretext task, (b) and then the pre-trained model is utilized to initialize the backbone and \emph{fine-tuned} with small-scale labeled or unlabeled person ReID datasets~(\eg, Market1501~\cite{zheng2015scalable}). In this paper, we focus on the first phase, \ieno, how to pre-train a ReID-friendly model in an unsupervised manner.


We first overview the whole pipeline of our UP-ReID in Section~\ref{section: overview}, and then introduce the proposed \emph{$I^2$}-regularization for pre-training, which comprises a global consistency constraint (Section~\ref{section: consistency constraint}) and an intrinsic contrastive constraint (Section~\ref{section: intrinsic contrastive contraint}). Last but not least, a prior-based hard mining strategy employed for local feature enhancement is discussed in Section~\ref{section: hard mining}.

\subsection{Overview}
\label{section: overview}

As illustrated in Figure~\ref{fig:Framework}, UP-ReID has two encoders: an online encoder $f_q$ and a momentum-based moving averaging (EMA) update encoder $f_k$. Both $f_q$ and $f_k$ are composed of a feature encoder and a projection head. The feature encoder is the model to be pre-trained~(\eg, ResNet50), and the projection head is a multi-layer perceptron. The online encoder $f_q$ will be updated by back-propagation and the EMA encoder $f_k$ will be slowly progressed through momentum-based moving average of the online encoder $f_q$, which is $\theta_k \gets m\theta_k + (1-m) \theta_q$. $\theta_k, \theta_q$ represent the parameters of $f_k, f_q$, and $m$ means the momentum coefficient.


Given an input image $x$, we can get two different views of $x$, \ie, a query view ${x}_{q,0}$ and a key view ${x}_{k,0}$, after two different augmentations. 
Unlike previous contrastive learning methods that only take the augmented images ${x}_{q,0}$ and ${x}_{k,0}$ as the input, we also feed the original image $x$ into the network as shown in Figure~\ref{fig:Framework}. Then, we enforce a consistency loss $\mathcal{L}_{consist.}$ to narrow down the distance between the similarity distribution of the augmented images and that of the original images in a mini-batch, which is described in detail in Section~\ref{section: consistency constraint}.

Moreover, before feeding ${x}_{q,0}$ and ${x}_{k,0}$ into the network, we partition each of them into $M$ non-overlapping patches. Note that, all $2M$ patches are partitioned from the same person image $x$ actually. Then, we feed these patches along with the entire augmented images into the online encoder and EMA encoder. An intrinsic contrastive loss $\mathcal{L}_{inc}$ is computed over them to learn both fine-grained local representations and the semantic image-level representations, which is discussed in detail in Section~\ref{section: intrinsic contrastive contraint}. For a better fine-grained information exploration, a hard mining strategy is further introduced to the calculation of the intrinsic contrastive loss, which is presented in Section~\ref{section: hard mining}. Ultimately, the total optimization objective is defined as:
\begin{equation}
\label{total loss}
    \mathcal{L}_{total} = \mathcal{L}_{consist.} + \mathcal{L}_{inc}.
\end{equation}

Additionally, a dynamic queue bank is constructed to store the feature representations of previous mini-batches and provide sufficient negative samples for the current mini-batch training. In practice, we prepare a queue for the image-level features, \ie, ${Q}_{0}$, and a queue for each patch-level local features, \ie, ${Q}_{i}$, $i\in\{1,...,M\}$. All of these queues together constitute the queue bank and they are dynamically updated by the features extracted by the EMA encoder.


\subsection{Consistency over Augmented-Original Images}
\label{section: consistency constraint}

Data augmentation plays a crucial role in contrastive learning. However, discriminative attributes of pedestrian images are very likely to be ruined by various augmentation operations (see Figure~\ref{fig:augmentation}). Due to the visual distortions caused by augmentation, a sample may be less similar to its positive instances but more similar to its negative samples instead, which inevitably imposes a negative effect on the pre-training process. 


To alleviate this problem, we turn to the original images for help. Although the identity-related features are possibly destroyed in the augmented images, those discriminative clues still remain in the original person images, \ie, raw images before augmentation. Thus, we propose to use the similarity between the original images as ground truth to supervise the images that go through the data augmentation, \ie, maintain the consistency before and after data augmentations.

For a mini-batch of input person images $x_{r}$, we get two groups of images $x_q$ and $x_k$ after two different data augmentations. Then we feed them into the network and we get the online encoder features $q$ and EMA encoder features $k$ respectively, that is, $q=f_q(x_q)$ and $k=f_k(x_k)$. The similarity distribution is computed as:
\begin{equation}
\label{eqn: APD}
    A(q,k) = \bm{q}\cdot\bm{k^T},
\end{equation}
where $q$ and $k$ have been normalized by the normalization layer followed by the projection head. $A(\cdot)$ denotes the inter-instance similarity calculation function between two batches of images after two different kinds of augmentation.

Similarly, we perform the same technique to the original input images $x_{r}$, which is expressed as $q_{r}=f_q(x_{r})$ and $k_{r}=f_k(x_{r})$. Then, we calculate the inter-instance similarity  distribution over the original images:
\begin{equation}
\label{eqn: OPD}
    A(q_r, k_r) = \bm{{q}_{r}}\cdot\bm{{k}^T_{r}}.
\end{equation}

After that, we employ a Maximum Mean Discrepancy (MMD)~\cite{gretton2012kernel} metric to measure the difference between two distributions and construct a consistency loss based on it:
\begin{equation}
\label{eqn:MMD}
    \mathcal{L}_{consist.} = MMD\left(A(q,k), \hspace{1mm}A(q_r, k_r)\right).
\end{equation}
Note that the calculated similarity distribution over the original images $A(q_r, k_r)$ just serves as the ground truth to supervise those with augmentations and does not participate in the update. So, there is no gradient back-propagation for the features of the original images. The consistency loss $\mathcal{L}_{consist.}$ helps the model to deduce and restore the discriminative local regions that are distorted by data augmentations, and further encourages the model to learn discriminative feature representations between different instances.

\subsection{Intrinsic Contrastive Constraint}
\label{section: intrinsic contrastive contraint}


To explore the intrinsic properties of a person image, we also introduce an intrinsic contrastive constraint in our UP-ReID framework. Before feeding the augmented images of $x_{q,0}$ and $x_{k,0}$ into the network (here we use subscript `0' to denote the holistic person image), we partition each of them into $M$ non-overlapping patches uniformly,
\begin{equation}
    \mathcal\{{x}_{q,1}, ..., {x}_{q,M}\} = P(x_{q,0}),
    \label{AP for query}
\end{equation}
\vspace{-5mm}
\begin{equation}
    \mathcal\{{x}_{k,1}, ..., {x}_{k,M}\} = P(x_{k,0}),
    \label{AP for key}
\end{equation}
where $P$ represents the partition operation, $x_{q,i}$ denotes the $i$-th patch partitioned from $x_{q,0}$, and $x_{k,i}$ denotes the $i$-th patch partitioned from $x_{k,0}$. Then, we group them together and get two sets:
$\mathcal{X}_{q}=\{{x}_{q,i}\}^M_{i=0}$ and $\mathcal{X}_{k}=\{{x}_{k,i}\}^M_{i=0}$.

Taking the image set $\mathcal{X}_{q}$ as an example for illustration, $\mathcal{X}_{q}$ comprises an image-level holistic instance $x_{q,0}$ and $M$ patch-level local instances $x_{q,i}$ ($i \in \{1,...,M\}$). All of them come from the input image $x$ and belong to the same instance, \ie, the input $x$. In short, $x_{q,0}$ contains the image-level global information while $x_{q,i}$ ($i \in \{1,...,M\}$) highlights the local information. 


As shown in Figure~\ref{fig:Framework}, we feed $\mathcal{X}_{q}$ and $\mathcal{X}_{k}$ into the online encoder $f_q$ and EMA encoder $f_k$, respectively, \ieno, $q_i=f_q(x_{q,i})$ and $k^+_i=f_k(x_{k,i})$, $i \in {0,1,...,M}$. To learn semantic-aware representations from the holistic images, we enforce a InfoNCE~\cite{oord2018representation} loss over the global features, which is formulated as:
\begin{equation}
\label{global contrastive loss}
    \mathcal{L}_{g} = - \text{log}\frac{\text{exp}(\bm{q_0}\cdot\bm{k^+_0} / \tau_1)}{\text{exp}(\bm{q_0}\cdot\bm{k^+_0} / \tau_1) + \sum_{j=0}^{N-1} \text{exp}(\bm{q_0}\cdot\bm{k^-_{0,j}/\tau_1})},
\end{equation}
where $\tau_1$ is the temperature hyper-parameter, $k^-_{0,j}$ is the negative sample in the image-level feature queue $Q_0$, and $N$ is the total number of negative samples in $Q_0$.

For the local fine-grained representation learning, we calculate a patch-wise contrastive loss over the patch-level instances. For the feature $q_i$, we denote its positive sample as $k^+_p$ and negative queue as $Q_n$. Formally, the patch-wise contrastive loss for the $i$-th patch $p_i$ is defined as:
\begin{equation}
\label{eqn:contrastive loss for patch i}
    \mathcal{L}_{p_i} = - \text{log}\frac{\text{exp}(\bm{q_i}\cdot\bm{k^+_p} / \tau_2)}{\text{exp}(\bm{q_i}\cdot\bm{k^+_p} / \tau_2) + \sum_{j=0}^{N-1} \text{exp}(\bm{q_i}\cdot\bm{k^-_{n,j}/\tau_2})},
\end{equation}
where $k^-_{n,j}$ is the negative sample in $Q_n$, and $\tau_2$ is the temperature hyper-parameter. The details about the selection of $k^+_p$ and $Q_n$ will be described in Section~\ref{section: hard mining}.

In order to fully explore the discriminative information contained in each body part of a pedestrian, we compute the aforementioned contrastive loss for each patch-level feature and take the weighted average sum of them as the final constraint. That is, the intrinsic contrastive loss is a weighted sum of $\mathcal{L}_{g}$ and multiple $\mathcal{L}_{p_i}$:
\begin{equation}
\label{identity-wise loss}
    \mathcal{L}_{inc} = \lambda_g * \mathcal{L}_{g} + \lambda_p * \frac{1}{M} \sum_{i=1}^{M} \mathcal{L}_{p_i},
\end{equation}
where $\lambda_g$ and $\lambda_p$ are the weighting parameters.

\subsection{Hard Mining for Local Feature Exploration}
\label{section: hard mining}

\begin{figure}[t]
    \centering
    \includegraphics[width=8.2cm, height=7.2cm]{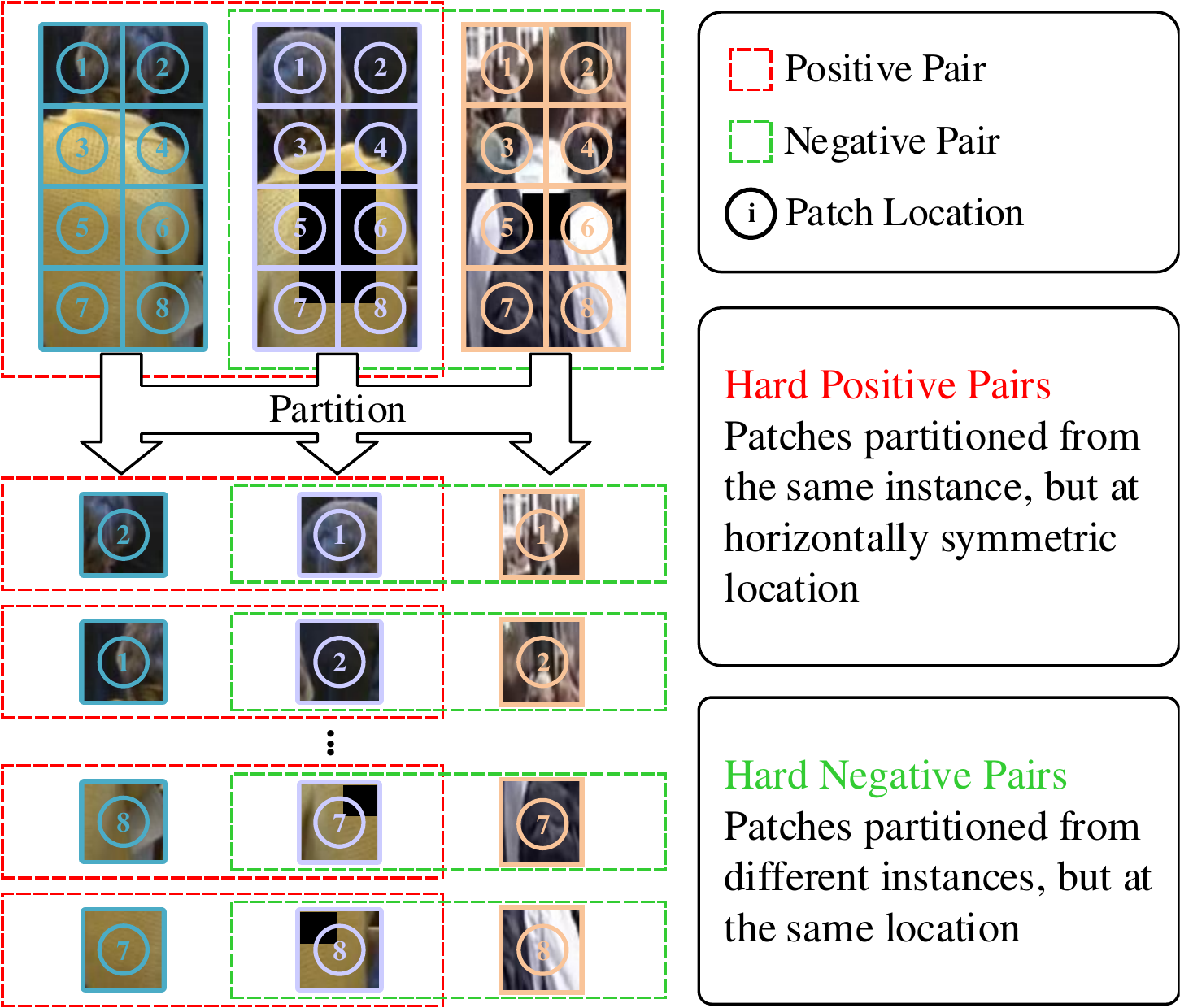}
    \setlength{\abovecaptionskip}{7pt}
    \vspace{-3mm}
    \caption{Illustration of our hard mining strategy. We choose two horizontally symmetric patches partitioned from the same instance as a positive pair, and two patches partitioned from different instances but at the same patch location as a negative pair.}
    \label{fig: hard mining}
    \vspace{-4mm}
\end{figure}


For the patch-level feature $q_i$ in Eq.~\ref{eqn:contrastive loss for patch i}, $k^+_i$ and $Q_i$ should be the positive sample and negative queue, respectively, \ie, $k^+_p=k^+_i,Q_n=Q_i$, corresponding to same patch region as $q_i$. For better representation learning, based on the prior knowledge that human body is horizontally symmetric, we further develop an effective hard mining method to select the positive sample and negative queue for each patch-level feature, which is shown in Figure~\ref{fig: hard mining}. 

\noindent\textbf{Hard Negative Queue Selection.}
The same body part of different persons could be discriminative, such as hair color and shoes color. Hence, for $q_i$ in Eq.~\ref{eqn:contrastive loss for patch i}, we choose the patches partitioned from different instances but at the same location as the negative samples (\ieno, $Q_n=Q_i$).

\noindent\textbf{Hard Positive Sample Selection.}
Considering the prior knowledge that persons are horizontally symmetric, we choose two horizontally symmetric patches partitioned from the same instance as a positive pair. 
Specifically, in Eq.~\ref{eqn:contrastive loss for patch i}, we select the feature $k^+_{i\_hs}$ (\ieno, the horizontally symmetrical patch feature corresponding to position $i$) as the positive sample of $q_i$.

Intuitively, the human body structure and clothing are mostly horizontally symmetric, which indicates that two symmetric patches of the same person image contain very similar visual representative patterns~(\eg, color, texture). This is important for person ReID. Thus, choosing them as a positive pair to pre-train the model is reasonable.
On the other hand, due to the different capture environments caused by camera angles or human postures, a pedestrian image may not be completely symmetrical, which means two symmetric patches have similar primary visual information but there are still differences in details. Thus, choosing them as a positive pair can improve the model's ability to identify similar visual representation patterns under different situations, which further helps the model recognize the same pedestrian under different environments.


Given that there are still some extreme cases that are totally inconsistent with the prior knowledge of horizontal symmetry of the pedestrian pictures~(\eg, pedestrian pictures taken from the side), we also select the same position patch of the other view person image (\ie, $k^+_i$) as one of the positive samples of $q_i$. So, the patch-wise contrastive loss in Eq.~\ref{eqn:contrastive loss for patch i} is modified to:

\begin{equation}
\footnotesize
\label{eqn:contrastive loss for patch i with hard mining}
    \mathcal{L}_{p_i} = - \text{log}\frac{\sum_{k^+_p\in\mathcal{P}(i)}\text{exp}(\bm{q_i}\cdot\bm{k^+_p} / \tau_2)}{\sum_{k^+_p\in\mathcal{P}(i)}\text{exp}(\bm{q_i}\cdot\bm{k^+_p} / \tau_2) + \sum_{j=0}^{N-1} \text{exp}(\bm{q_i}\cdot\bm{k^-_{i,j}/\tau_2})},
\end{equation}
where $\mathcal{P}(i)=\{k^+_{i\_hs}, k^+_i\}$, $k^-_{i,j}\in Q_i$. 




\section{Experiments}

\subsection{Implementation}

\noindent\textbf{Training details.} 
For fair comparison, we use ResNet50 as the pre-trained backbone model and SGD as the optimizer.
The input images are resized to 256$\times$128. The mini-batch size is set to 800, and the initial learning rate is $0.1$. In our experiments, $M$ is set to $8$, $N$ is set to 65536, $m$ is set to 0.9, $\tau_1$ and $\tau_2$ are both set to 0.1, $\lambda_g$ and $\lambda_p$ are set to $0.8$ and $0.2$.  The pre-training models are trained with 8$\times$2080Ti GPUs for 3 weeks under Pytorch framework.

\noindent\textbf{Augmentation and Patch Partition.} 
Data augmentation plays a crucial role in self-supervised contrastive learning. We utilize the same augmentation operations as ~\cite{fu2021unsupervised}.
As for partition, we adopt image-level partition strategy. Specifically, we first partition a holistic image into multiple horizontal stripes, and then divide each stripe vertically into two patches uniformly.
It is necessary to emphasize that we apply the global-level augmentation (\ieno, augmentation followed by partition) rather than patch-level augmentation (\ieno, partition followed by augmentation). Because the global-level augmentation is closer to the realistic data variation and will not break the inherent consistency among patches partitioned from the same person image. 

\noindent\textbf{Datasets.} 
We pre-train our model on ``LUPerson"~\cite{fu2021unsupervised} dataset. To demonstrate the superiority of our pre-trained model, we conduct extensive downstream experiments on four public ReID datasets, including CUHK03~\cite{li2014deepreid}, Market1501~\cite{zheng2015scalable}, PersonX~\cite{sun2019dissecting}, and MSMT17~\cite{wei2018person}. Note that we do not use DukeMTMC~\cite{zheng2017unlabeled} to avoid ethical issues.

\noindent\textbf{Evaluation Protocols.} 
Following the standard evaluation metrics, we use the cumulative matching characteristics at Rank1 and mean average precision (mAP) to evaluate the performance.

\begin{table*}[t]
\caption{Comparison of three representative supervised ReID methods using different pre-trained models in terms of mAP/Rank1 (\%). ``INSUP" refers to the supervised pre-trained model on ImageNet, ``Moco" and ``UP-ReID" refer to the Moco and our UP-ReID pre-trained models on LUPerson, respectively. More comparison results can be found in \textbf{Appendix}.}
\setlength{\tabcolsep}{3.3mm}
    \begin{subtable}[h]{0.5\textwidth}
        \centering
        \begin{tabular}{l|ccc}
        \shline
        Model & BDB~\cite{dai2019batch} & BOT~\cite{luo2019bag} & MGN~\cite{wang2018learning} \\
        \hline
        INSUP    & 76.7/79.4 & 62.0/63.9 & 70.5/71.2 \\ \hline
        Moco  & 78.9/81.5 & 66.7/66.3 & 74.7/75.4 \\ \hline
        UP-ReID & \textbf{79.6/82.6} & \textbf{68.7/69.1} & \textbf{85.3/87.6} \\ \shline
        \end{tabular}
        \caption{CUHK03}
        \label{tab:improve-cuhk}
    \end{subtable}
    \hfill
    \begin{subtable}[h]{0.5\textwidth}
    \centering
        \begin{tabular}{l|ccc}
        \shline
        Model  & BDB~\cite{dai2019batch} & BOT~\cite{luo2019bag} & MGN~\cite{wang2018learning} \\
        \hline
        INSUP    & 86.7/95.3 & 85.7/94.3 & 87.5/95.1 \\ \hline
        Moco  & 88.1/95.3 & 87.6/94.9 & 91.0/96.4 \\ \hline
        UP-ReID & \textbf{88.5/95.3} & \textbf{88.1/95.1} & \textbf{91.1/97.1} \\ \shline
        \end{tabular}
        \caption{Market1501}
        \label{tab:improve-market}
    \end{subtable}
    \hfill
    \begin{subtable}[h]{0.5\textwidth}
    \centering
        \begin{tabular}{l|ccc}
        \shline
        Model & BDB~\cite{dai2019batch} & BOT~\cite{luo2019bag} & MGN~\cite{wang2018learning} \\
        \hline
        INSUP    & 84.4/95.1 & 86.7/94.8 & 85.3/94.3 \\ \hline
        Moco  & 84.8/95.2 & 86.5/94.6 & 85.8/94.2 \\ \hline
        UP-ReID & \textbf{86.1/95.3} & \textbf{88.0/95.3} & \textbf{89.7/96.1} \\ \shline
        \end{tabular}
        \caption{PersonX}
        \label{tab:improve-PersonX}
    \end{subtable}
    \hfill
    \begin{subtable}[h]{0.5\textwidth}
    \centering
        \begin{tabular}{l|ccc}
        \shline
        Model & BDB~\cite{dai2019batch} & BOT~\cite{luo2019bag} & MGN~\cite{wang2018learning} \\
        \hline
        INSUP    & 49.2/77.4 & 53.4/76.8 & 61.5/84.0 \\ \hline
        Moco  & 51.2/78.1 & 53.2/75.4 & 62.9/83.9 \\ \hline
        UP-ReID & \textbf{52.4/78.7} & \textbf{56.2/78.1} & \textbf{63.3/84.3} \\ \shline
        \end{tabular}
        \caption{MSMT17}
        \label{tab:improve-msmt}
    \end{subtable}
    \vspace{-2mm}
    \label{table1: supervised reid comparison}
\end{table*}


\begin{figure*}[h!]
\centering
\begin{subfigure}{0.33\linewidth}
    \includegraphics[width=1.0\linewidth]{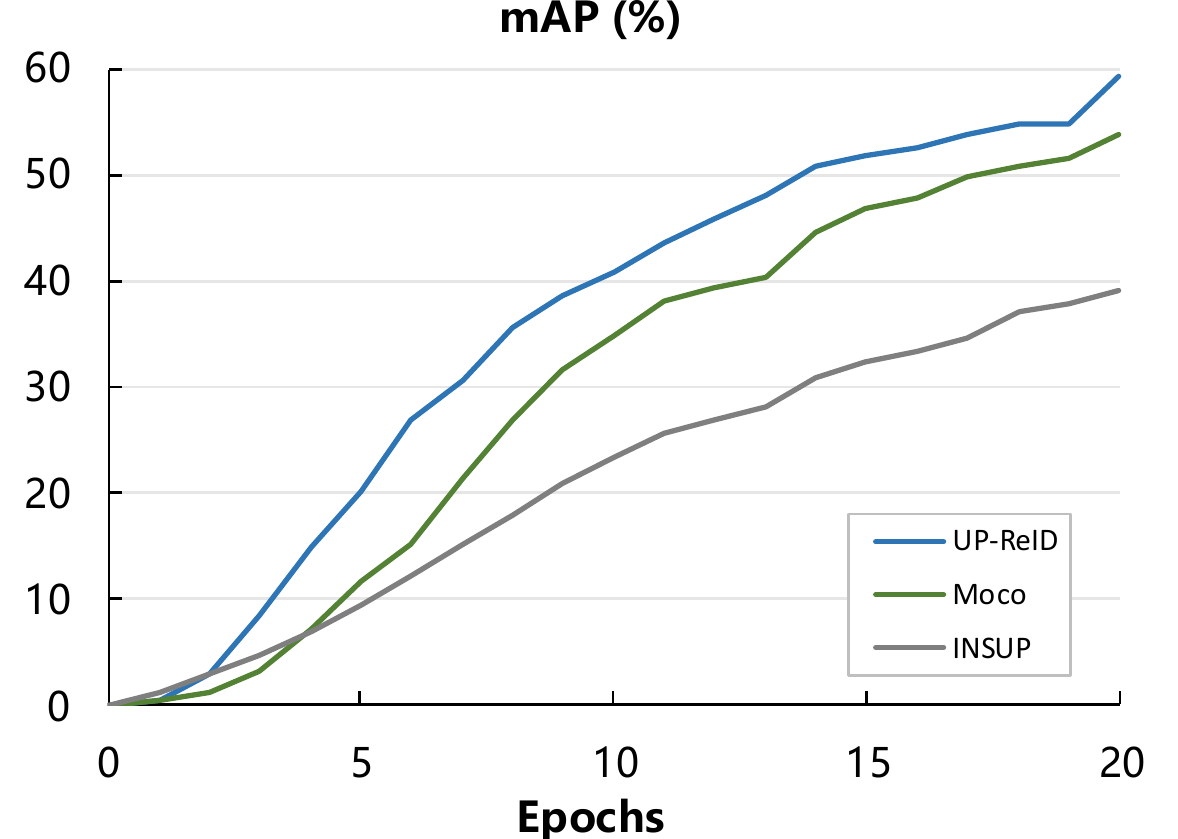}
    \caption{mAP learning curve on CUHK03}
    \label{subfig: convergence on cuhk03}
\end{subfigure}
\begin{subfigure}{0.33\linewidth}
    \includegraphics[width=1.0\linewidth]{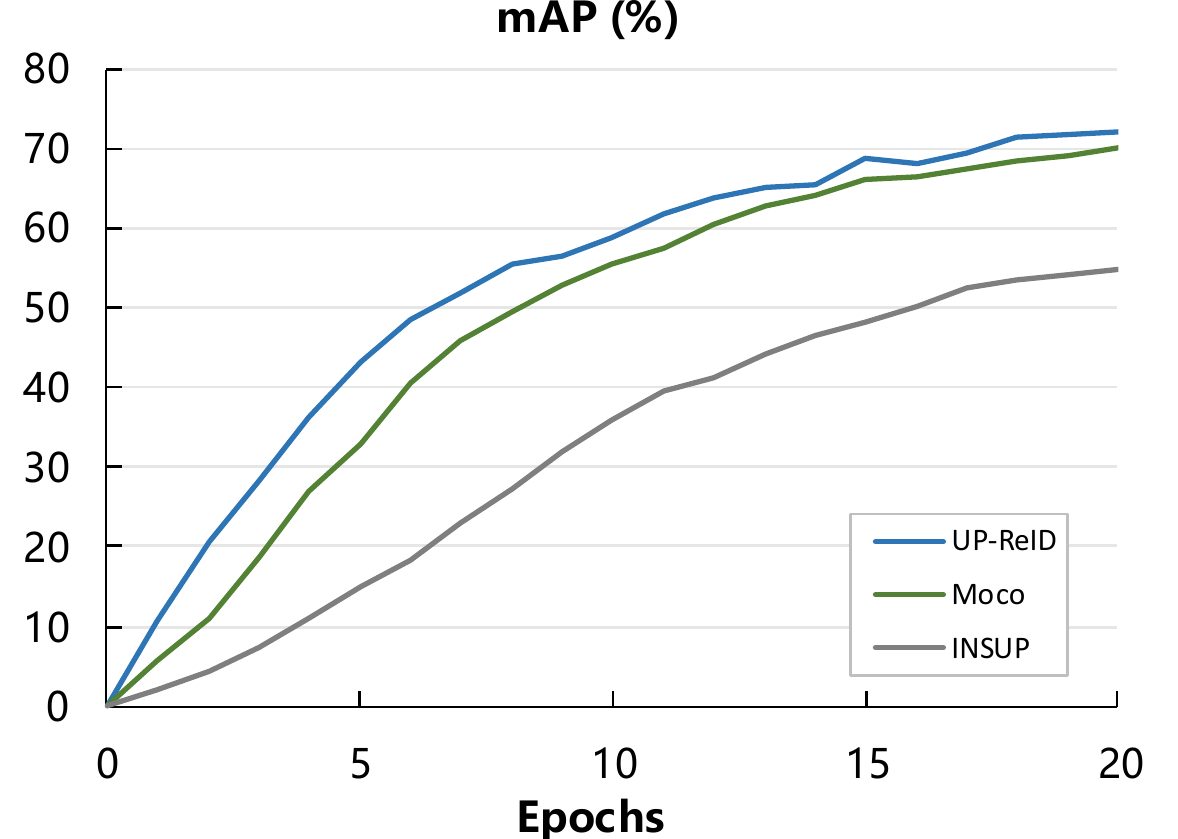}
    \caption{mAP learning curve on Market1501}
    \label{subfig: convergence on market1501}
\end{subfigure}
\begin{subfigure}{0.33\linewidth}
    \includegraphics[width=1.0\linewidth]{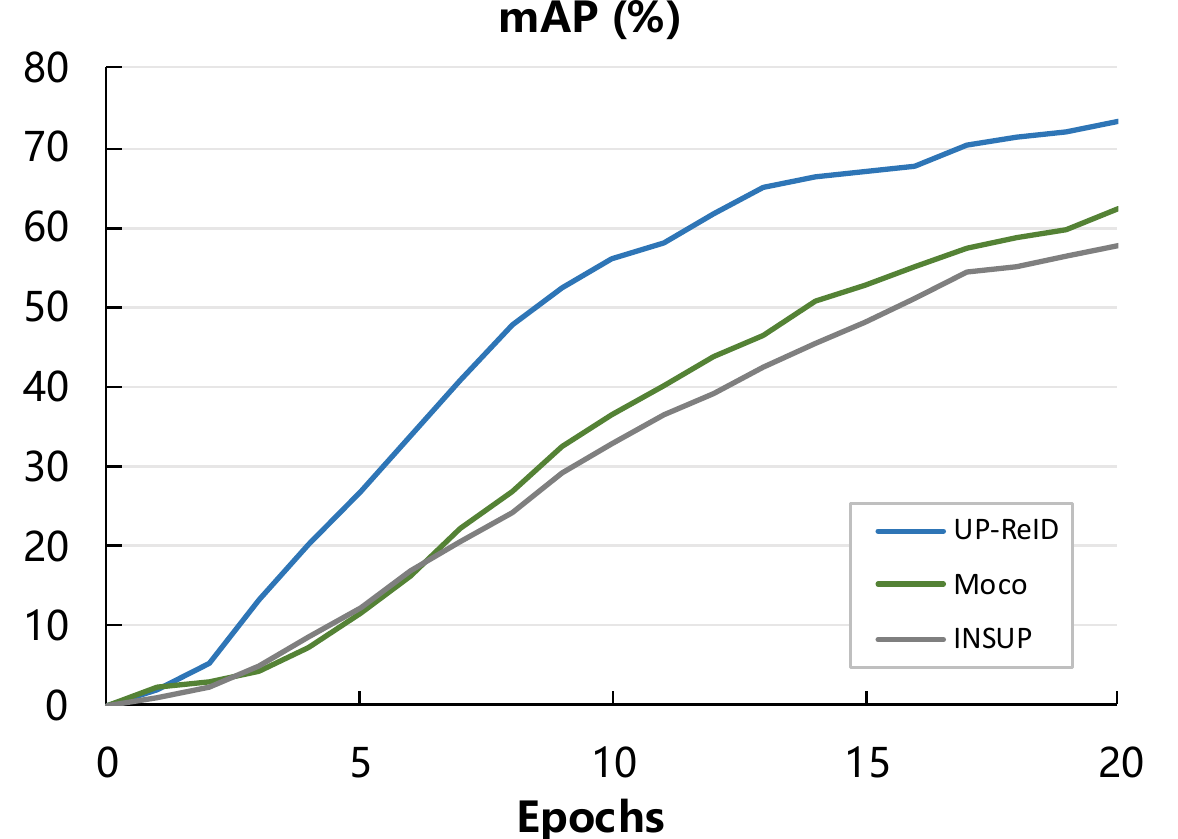}
    \caption{mAP learning curve on PersonX}
    \label{subfig: convergence on personx}
\end{subfigure}
\vspace{-2mm}
\caption{mAP learning curves with different pre-trained models in BDB on three datasets (CUHK03, Market1501, and PersonX) with the same training schedule. More comparison results can be found in \textbf{Appendix}.}
\label{fig: convergence rapidity}
\vspace{-2mm}
\end{figure*}


\subsection{Improvement on Supervised ReID}
In this section, we show the superiority of our UP-ReID by comparing with the model unsupervisedly pre-trained on LUPerson by Moco~\cite{fu2021unsupervised} and the commonly used supervised pre-trained model on ImageNet in three representative supervised ReID approaches: Batch DropBlock Network (BDB)~\cite{dai2019batch}, Strong Baseline (BOT)~\cite{luo2019bag} and Multiple Granularity Network (MGN)~\cite{wang2018learning}.
The BDB is re-implemented based on the open source code. As for BOT and MGN, we implement them in fast-reid~\cite{he2020fastreid}. 

Table~\ref{table1: supervised reid comparison} shows the improvements in the three selected supervised ReID methods on four popular person ReID datasets.
It can be seen that, compared to initializing with Moco, the MGN with UP-ReID has achieved \textbf{12.2\%}, \textbf{0.7\%}, \textbf{1.9\%}, \textbf{0.4\%} improvements in terms of Rank1 on CUHK03, Market1501, PersonX, MSMT17, respectively; BOT also has achieved \textbf{2.8\%}, \textbf{0.2\%}, \textbf{0.7\%}, \textbf{2.7\%} improvements in terms of Rank1 on these four datasets. 

Figure~\ref{fig: convergence rapidity} describes the comparison of the convergence speed of applying different pre-trained models in method BDB at the early stage of fine-tuning. UP-ReID outperforms both Moco and INSUP with faster convergence on all three datasets. The performance enhancement is more noticeable on PersonX (see Figure~\ref{subfig: convergence on personx}). On the Market1501 where the advantage is not obvious, UP-ReID still holds the lead of Moco by 1.7\% mAP improvement (see Figure~\ref{subfig: convergence on market1501}).

\subsection{Improvement on Unsupervised ReID}

Our pre-trained model can also benefit unsupervised ReID methods. To demonstrate this, we test our pre-trained model on SpCL~\cite{ge2020self}. We evaluate the performance on Market1501 and PersonX.

In Table~\ref{tab:improvement-unsup}, M means purely unsupervised training on Market1501, and P $\rightarrow$ M means unsupervised domain adaptation whose source dataset is PersonX and target dataset is Market1501. As we can see, UP-ReID outperforms Moco by \textbf{2.9\%}, \textbf{6.3\%} in terms of mAP and \textbf{2.2\%}, \textbf{2.5\%} in terms of Rank1 on M and P $\rightarrow$ M, respectively. It further verifies that UP-ReID can achieve better superiority and generalization capability for person ReID.
Note that we implement SpCL by official OpenUnReid~\cite{ge2020self}.

\vspace{-3mm}
\begin{table}[h]
\caption{Performance (\%) comparisons of using different pre-trained models on unsupervised ReID method SpCL.}
\vspace{-3mm}
    \centering
    \begin{tabular}{P{1.5cm}|C{0.8cm}C{0.8cm}|C{0.8cm}C{0.8cm}}
    \shline
    \centering
    \multirow{2}{*}{Model} & \multicolumn{2}{c|}{M} & \multicolumn{2}{c}{P $\rightarrow$ M} \\
    \cline{2-5} & mAP & Rank1 & mAP & Rank1 \\
    \hline
    \centering
    INSUP    & 73.1 & 88.1 & 73.8 & 88.0 \\
    \centering
    Moco  & 72.2 & 87.8 & 72.4 & 88.4 \\
    \centering
    UP-ReID & \textbf{75.1} & \textbf{90.0} & \textbf{78.7} & \textbf{90.9} \\
    \shline
\end{tabular}
\vspace{-1mm}
\centering
\label{tab:improvement-unsup}
\end{table}
\vspace{-5mm}

\subsection{Comparison with State-of-the-Art Methods}

In this section, we compare our results with state-of-the-art methods on CUHK03 and Market1501 datasets. Notice that we do not use any additional modules (\egno, IBN-Net) or post-processing methods (\egno, Re-Rank~\cite{zhong2017re}).We just simply apply UP-ReID pre-trained vanilla ResNet50 on MGN. As shown in Table~\ref{tab:sota}, MGN equipped with UP-ReID ResNet50 outperforms all compared methods on both datasets.

\begin{table}[h!]
\small
\centering
\caption{Performance (\%) comparisons with state-of-the-art approaches on CUHK03 and Market1501. The best results are marked as bold and the second ones are masked by underline. We show more comparison results in \textbf{Appendix}.}
\vspace{-2mm}
\begin{tabular}{P{3.50cm}|C{0.67cm}C{0.67cm}|C{0.67cm}C{0.67cm}}
    \shline
    \multirow{2}{*}{Methods} & \multicolumn{2}{c|}{CUHK03} & \multicolumn{2}{c}{Market1501} \\
    \cline{2-5} & mAP & Rank1 & mAP & Rank1 \\ 
    \hline
    PCB~\cite{sun2018beyond} (ECCV'18) & 57.5 & 63.7 & 81.6 & 93.8 \\
    OSNet~\cite{zhou2019omni} (ICCV'19) & 67.8 & 72.3 & 84.9 & 94.8 \\
    P2Net~\cite{guo2019beyond} (ICCV'19) & 73.6 & 78.3 & 85.6 & 95.2 \\
    SCAL~\cite{chen2019self} (ICCV'19) & 72.3 & 74.8 & 89.3 & 95.8 \\
    DSA~\cite{zhang2019densely} (CVPR'19) & 75.2 & 78.9 & 87.6 & 95.7 \\
    GCP~\cite{park2020relation} (AAAI'20) & 75.6 & 77.9 & 88.9 & 95.2 \\
    SAN~\cite{jin2020semantics} (AAAI'20) & 76.4 & 80.1 & 88.0 & \underline{96.1} \\
    ISP~\cite{zhu2020identity} (ECCV'20) & 74.1 & 76.5 & 88.6 & 95.3 \\
    GASM~\cite{he2020guided} (ECCV'20) & - & - & 84.7 & 95.3 \\
    RGA-SC~\cite{zhang2020relation} (CVPR'20) & \underline{77.4} & \underline{81.1} & 88.4 & \underline{96.1} \\
    HOReID~\cite{wang2020high} (CVPR'20) & - & - & 84.9 & 94.2 \\
    AMD~\cite{chen2021explainable} (ICCV'21) & - & - & 87.1 & 94.8 \\
    TransReID~\cite{he2021transreid} (ICCV'21) & - & - & \underline{89.5} & 95.2  \\
    PAT~\cite{li2021diverse} (CVPR'21) & - & - & 88.0 & 95.4 \\
    \hline
    MGN+UP-ReID (Ours) & \textbf{85.3} & \textbf{87.6} & \textbf{91.1} & \textbf{97.1} \\
    \shline
\end{tabular}\\
\label{tab:sota}
\end{table}

\subsection{Ablation Study}
In this section, we perform comprehensive ablation studies to demonstrate the effectiveness of our designs in the proposed UP-ReID. Here we fine-tune different pre-trained models with supervised ReID method MGN~\cite{wang2018learning} on CUHK03 to validate the effectiveness of each component. 

\noindent\textbf{Effectiveness of the Consistency Constraint and the Intrinsic Contrastive Constraint.}
Our UP-ReID consists of two key constraints: the consistency constraint (CC) and the intrinsic contrastive constraint (ICC). We evaluate the benefits of them in Table~\ref{tab: effectiveness of CC and ICC.}. Specifically, (b) Baseline with CC and (c) Baseline with ICC outperform the (a) Baseline by \textbf{4.4\%/4.8\%} and \textbf{6.7\%/8.2\%} in terms of mAP/Rank1 on CUHK03, respectively. With both two constraints, (d) UP-ReID achieves \textbf{85.3\%(+10.6\%)} mAP and \textbf{87.6\%(+12.2\%)} Rank1 on CUHK03, which demonstrates that CC and ICC are complementary and both vital to UP-ReID, jointly resulting in a superior performance.

\begin{figure}[h!]
    \centering
    \includegraphics[width=8.5cm, height=6.1cm]{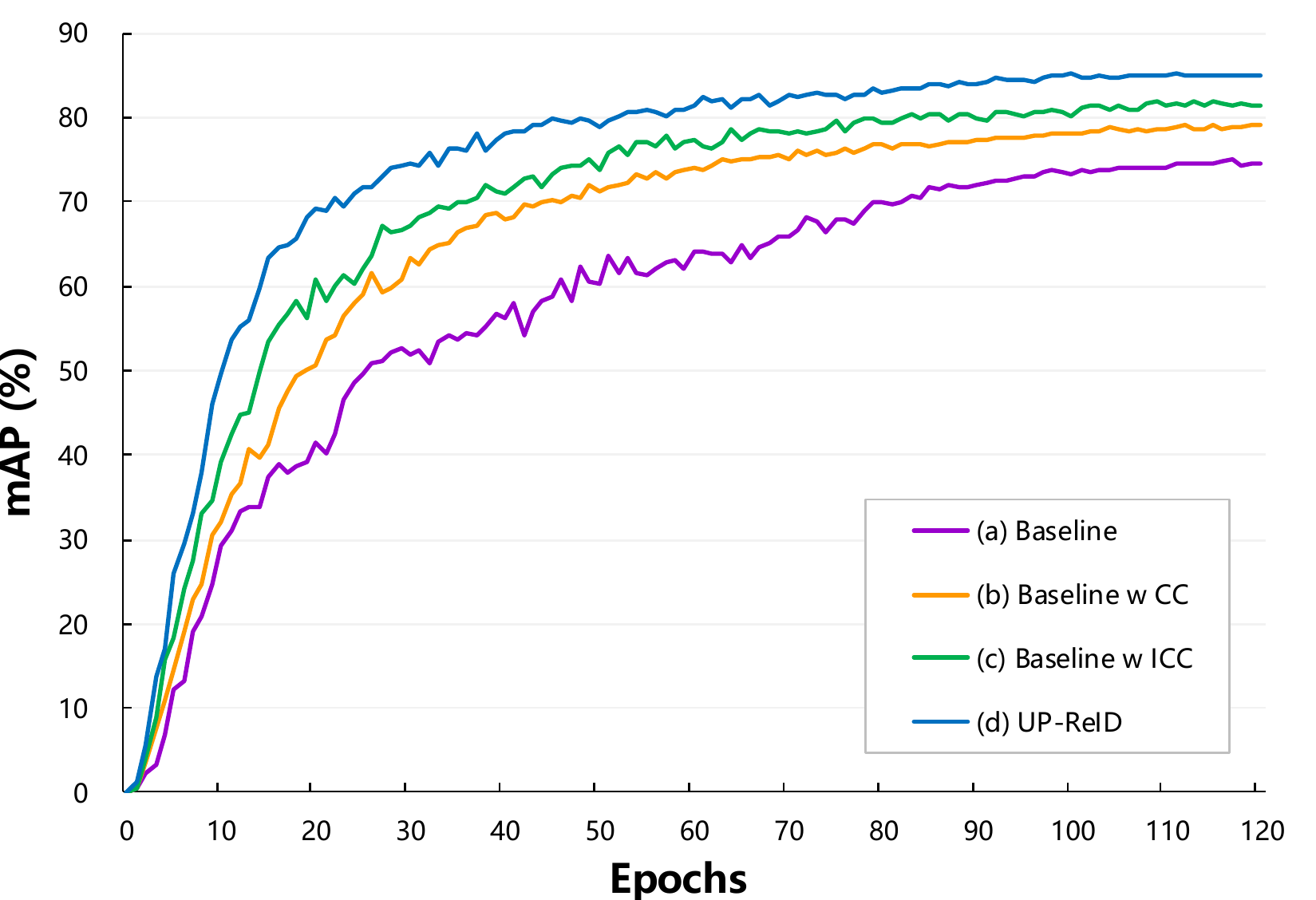}
    \setlength{\abovecaptionskip}{7pt}
    \vspace{-5mm}
    \caption{mAP learning curves of CUHK03 in MGN with four different pre-trained UP-ReID models. The models correspond to the models in Table~\ref{tab: effectiveness of CC and ICC.} one-to-one.}
    \label{fig: ablation}
\end{figure}

We also evaluate the effectiveness of each components of our UP-ReID in terms of the convergence speed on CUHK03.
Figure~\ref{fig: ablation} plots the mAP learning curves of four different pre-trained models with MGN. As we can see, the (b) Baseline with CC and (c) Baseline with ICC achieve faster convergence than (a) Baseline. More importantly, (d) the UP-ReID with both constraints (\ieno, ICC and CC) achieves faster convergence than both (b) and (c) which only have one constraint.


The experimental results demonstrate that both the consistency constraint and the intrinsic contrastive constraint contribute to a better visual representation. The former is designed to counter the augmentation perturbations, and the latter is designed for detailed information exploration.

\vspace{-2mm}
\begin{table}[h!]
\caption{The ablation results of several variants of UP-ReID pre-trained models that are fine-tuned on CUHK03. The values in the brackets are the improvement compared to the Baseline.}
\vspace{-2mm}
\small
    \centering
    \begin{tabular}{c|cc|cc}
        \shline
        Model & CC & ICC & mAP & Rank1 \\
        \hline
        (a) Baseline & $\times$ & $\times$ & 74.7 & 75.4 \\
        (b) Baseline w CC & $\surd$ & $\times$ & 79.1(+4.4) & 80.2(+4.8) \\
        (c) Baseline w ICC & $\times$ & $\surd$ & 81.4(+6.7) & 83.6(+8.2) \\
        (d) UP-ReID & $\surd$ & $\surd$ & \textbf{85.3}(+10.6) & \textbf{87.6}(+12.2) \\
        \shline
    \end{tabular}
    \label{tab: effectiveness of CC and ICC.}
\end{table}

\vspace{-2mm}

\noindent\textbf{Effectiveness of the Hard Mining Strategy.} For better representation learning, we introduce a hard mining (HM) strategy to the intrinsic contrastive constraint. As shown in Table~\ref{tab: effectiveness of Hard Mining.}, the UP-ReID without hard mining strategy (\ie, replace Eq.~\ref{eqn:contrastive loss for patch i with hard mining} with Eq.~\ref{eqn:contrastive loss for patch i}) has a 4.6\%/4.5\% drop in mAP/Rank1. Obviously, our hard mining strategy improves the discrimination and generalization capability of the pre-trained model.

Different from the previous works~\cite{hermans2017defense}, we select positive and negative pairs based on the prior knowledge that persons are horizontally symmetric instead of an online way. We further investigate the influence of different hard mining strategies and show more results in \textbf{Appendix}.

\begin{table}[h!]
\caption{Effectiveness of the hard mining strategy for ICC in our UP-ReID on CUHK03.}
    \centering
    \begin{tabular}{c|ccc}
        \shline
        Model & mAP & Rank1 & Rank5 \\
        \hline
        UP-ReID w/o HM & 80.7 & 83.1 & 93.1 \\
        UP-ReID w HM & \textbf{85.3} & \textbf{87.6} & \textbf{95.4} \\
        \shline
    \end{tabular}
    \vspace{-4mm}
    \label{tab: effectiveness of Hard Mining.}
\end{table}

\noindent\textbf{Influence of the Number of the Patch-Level Instances.}
Note that each patch-level instance is partitioned from the corresponding image-level instance. Different patch number ($M$) means different patch size. We investigate the influence of the patch-level instance number in the intrinsic contrastive constraint. 
As described in Table~\ref{tab: different number patches.}, $M$=8 outperforms $M$=4 by \textbf{4.0\%/4.5\%} in mAP/Rank1 on CUHK03, which also surpasses $M$=12 by \textbf{4.6\%/5.4\%} in mAP/Rank1. When $M$=8, each patch-level instance has a proper size, which is neither too large to ignore discriminative attributes, nor too small to introduce unnecessary noise.

\vspace{-1mm}
\begin{table}[h!]
\caption{Results of different number of patches in ICC.}
\vspace{-1mm}
    \centering
    \begin{tabular}{c|ccc}
        \shline
        Model & mAP & Rank1 & Rank5 \\
        \hline
        UP-ReID w $M=4$ & 81.3 & 83.1 & 92.6 \\
        UP-ReID w $M=12$ & 80.7 & 82.2 & 92.4 \\
        UP-ReID w $M=8$ & \textbf{85.3} & \textbf{87.6} & \textbf{95.4} \\
        \shline
    \end{tabular}
    \label{tab: different number patches.}
    \vspace{-5mm}
\end{table}

\section{Conclusion}
In this paper, we present two critical issues in applying contrastive learning to ReID pre-training task. Then, we propose a ReID-specific pre-training framework UP-ReID with an intra-identity regularization, which consists of a global consistency constraint and an intrinsic contrastive constraint. Moreover, we introduce a hard mining strategy to local information exploration for better representation learning. Extensive experiments demonstrate that UP-ReID could improve the downstream works performance with higher precision and much faster convergence. We hope more methods can be motivated such as unsupervised pre-training for ReID-specific transformers and apply UP-ReID to more downstream tasks~(\eg, occluded person ReID).

{\small
\bibliographystyle{ieee_fullname}
\bibliography{egbib}
}

\newpage
\appendix

\noindent\textbf{\Large Appendix}

\section{Datasets}
\subsection{Pre-training Dataset}
\noindent\textbf{LUPerson}~\cite{fu2021unsupervised} consists of 4,180,243 person images of over 200K identities extracted from 46,260 YouTube videos. YOLO-v5 trained on MS-COCO is utilized to extract each person instance in the sampled frame. It is worth noting that the LUPerson is large enough to support unsupervised person ReID feature learning.

\subsection{Fine-tuning Datasets}
\noindent\textbf{CUHK03}~\cite{li2014deepreid} contains 13,164 images of 1,360 pedestrians. Each identity is observed by 2 cameras. Note that CUHK03 offers both hand-labeled and DPM-detected bounding boxes, and the former is adopted in this paper.

\noindent\textbf{Market1501}~\cite{zheng2015scalable} contains 32,668 person images of 1,501 identities captured by 6 cameras. The training set consists of 12,936 images of 751 identities, the query set consists of 3,368 images, and the gallery set consists of 19,732 images of 750 identities.

\noindent\textbf{PersonX}~\cite{sun2019dissecting} is a large-scale data synthesis engine, which contains 1,266 manually designed identities and editable visual variables. Each identity is captured by 6 cameras.

\noindent\textbf{MSMT17}~\cite{wei2018person} contains of 126,441 images of 4,101 identities captured by 15 cameras. The training set consists of 30,248 person images of 1,041 identities, the query set consists of 11,659 images, and the gallery consists of 82,161 images of 3,060 identities.

\section{More Details about Data Augmentation}
Data augmentation plays a crucial role in self-supervised contrastive learning. We adopt popular augmentation operations including resizing, cropping, random grayscale, Gaussian blurring, horizontal flipping, and RandomErasing. Note that we abandon color jitter since person ReID is extremely dependent on color information~\cite{fu2021unsupervised}.

\section{Additional Results}

\subsection{More Results for Supervised ReID}
In Section 4.2 of the main body, we demonstrate that our UP-ReID can benefit the supervised ReID methods and show the results in Table 1. Here, we present the remaining results. Table~\ref{tab:comparison on PCB} shows the results of using different pre-trained models in the supervised fine-tuning ReID method PCB~\cite{sun2018beyond} on CUHK03, Market1501, and PersonX.

\begin{table}[h!]
	\centering
	\caption{Comparison of PCB method using different pre-trained models on three datasets in terms of mAP/Rank1 (\%).}
	\begin{tabular}{l|ccc}
		\shline
		Model & CUHK03 & Market1501 & PersonX \\
		\hline
		INSUP    & 59.5/69.9 & 78.0/92.6 & 80.9/92.7 \\ \hline
		Moco  & 58.3/72.8 & 79.3/92.9 & 80.7/92.9 \\ \hline
		UP-ReID & \textbf{60.1/74.1} & \textbf{80.0/93.1} & \textbf{81.7/93.2} \\ \shline
	\end{tabular}
	\label{tab:comparison on PCB}
\end{table}

\begin{figure*}[t!]
	\centering
	\begin{subfigure}{0.33\linewidth}
		\includegraphics[width=1.0\linewidth]{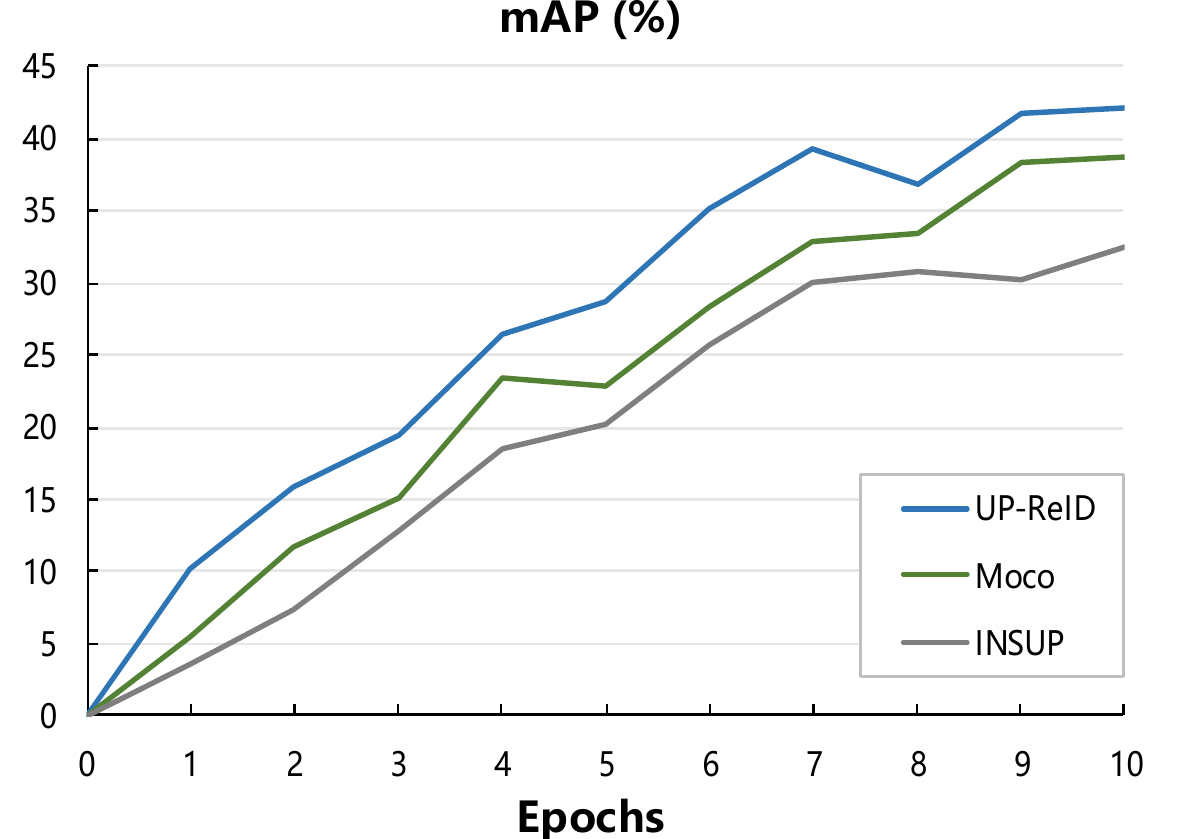}
		\caption{mAP learning curve on CUHK03}
		\label{subfig: convergence on cuhk03}
	\end{subfigure}
	\begin{subfigure}{0.33\linewidth}
		\includegraphics[width=1.0\linewidth]{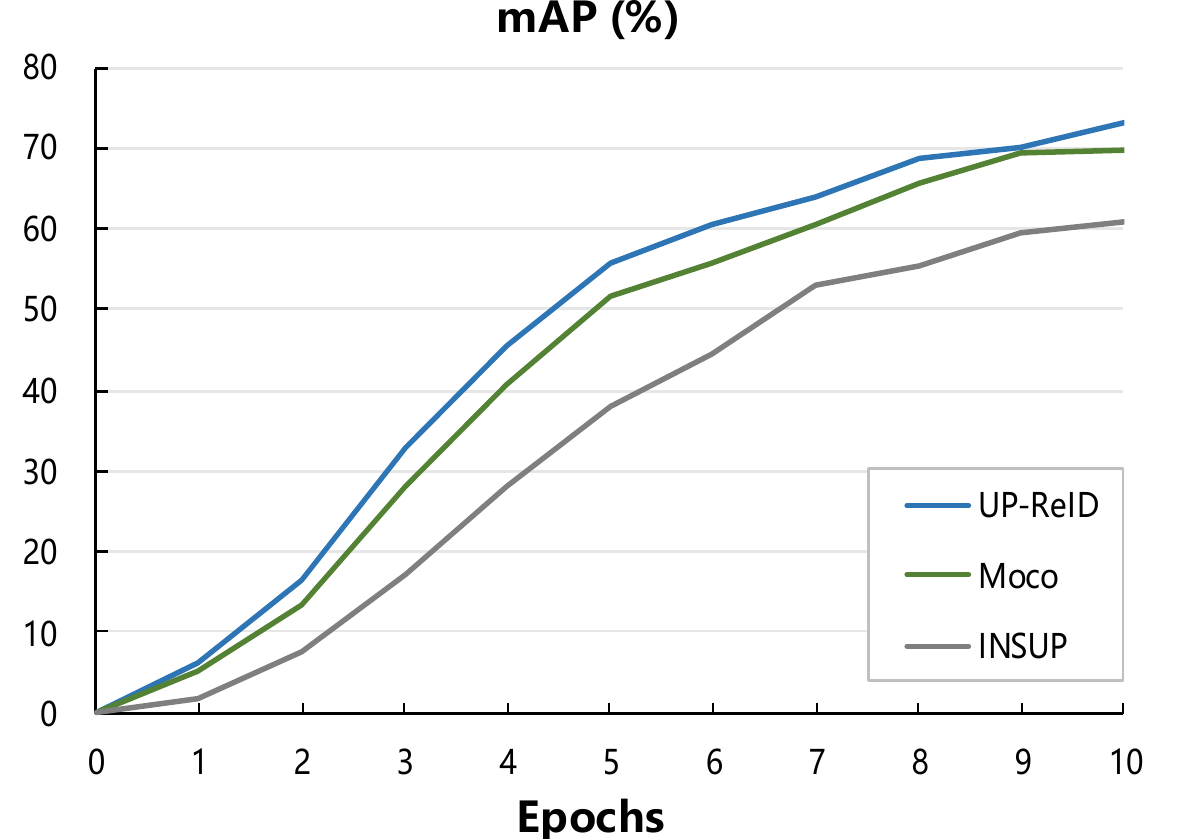}
		\caption{mAP learning curve on Market1501}
		\label{subfig: convergence on market1501}
	\end{subfigure}
	\begin{subfigure}{0.33\linewidth}
		\includegraphics[width=1.0\linewidth]{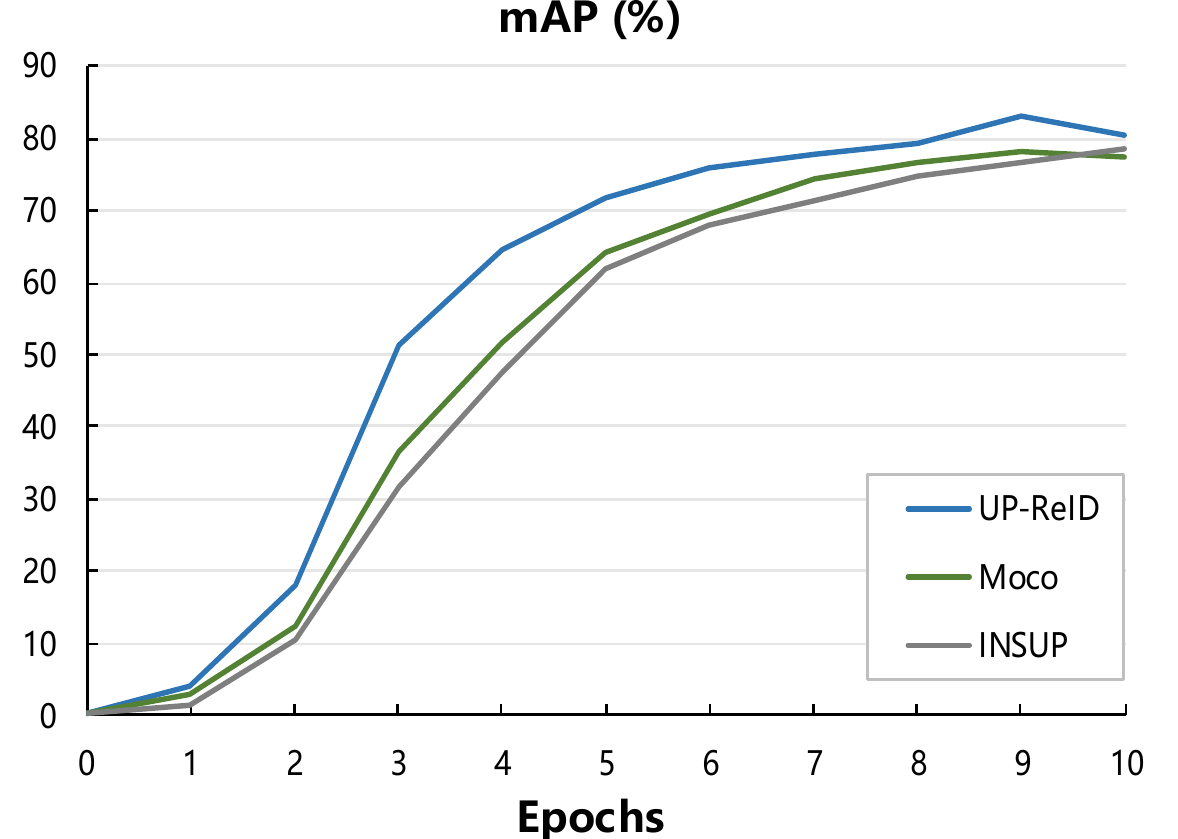}
		\caption{mAP learning curve on PersonX}
		\label{subfig: convergence on personx}
	\end{subfigure}
	\vspace{-2mm}
	\caption{mAP learning curves of different pre-trained models in MGN~\cite{wang2018learning} on three datasets (CUHK03, Market1501, and PersonX) with the same training schedule.}
	\label{fig: convergence rapidity in MGN}
\end{figure*}

We also show the comparison of the convergence speed of applying different pre-trained models in method MGN~\cite{wang2018learning} at the early stage of fine-tuning in Figure~\ref{fig: convergence rapidity in MGN}. As can be seen, UP-ReID achieves a faster convergence rapidity compared with Moco and INSUP on all the three datasets, which further demonstrates that the proposed UP-ReID can better benefit downstream ReID tasks.

\subsection{More Comparisons with State-of-the-Arts}
In Section 4.4 of the main body, we have shown some comparison results between our UP-ReID and state-of-the-art methods. Here, we extend the results in Table 3 and show the complete results of the comparison between UP-ReID and state-of-the-art methods in Table~\ref{tab:sota2} on three datasets, including CUHK03, Market1501, and MSMT17.
As we can see, MGN with our UP-ReID outperforms the other methods by at least \textbf{7.9\%/6.5\%} and \textbf{1.6\%/1.0\%} in terms of mAP/Rank1 on CUHK03 and Market1501, respectively. On the MSMT17 dataset, the TransReID~\cite{he2021transreid} achieves better performance. However, TransReID adopts transformer-based network and utilizes camera information additionally.

\begin{table*}[t!]
	\centering
	\caption{Complete performance (\%) comparisons with state-of-the-art approaches on CUHK03, Market1501, and MSMT17. The best results are marked as bold and the second ones are masked by underline.}
	\begin{tabular}{P{4.5cm}|C{1.4cm}C{1.4cm}|C{1.4cm}C{1.4cm}|C{1.4cm}C{1.4cm}}
		\shline
		\centering
		\multirow{2}{*}{Method} & \multicolumn{2}{c|}{CUHK03} & \multicolumn{2}{c|}{Market1501} & \multicolumn{2}{c}{MSMT17}\\
		\cline{2-7}
		\multicolumn{1}{c|}{} & mAP & cmc1 & mAP & cmc1 & mAP & cmc1 \\ 
		\hline 
		PCB~\cite{sun2018beyond} (ECCV'18) & 57.5 & 63.7 & 81.6 & 93.8 & - & - \\
		MGN~\cite{wang2018learning} (ACM MM'18) & 70.5 & 71.2 & 86.9 & 95.7 & - & - \\
		ABDNet~\cite{chen2019abd} (ICCV'19) & - & - & 88.3 & 95.6 & 60.8 & 82.3 \\
		BDB~\cite{dai2019batch} (ICCV'19) & 76.7 & 79.4 & 86.7 & 95.3 & - & - \\
		OSNet~\cite{zhou2019omni} (ICCV'19) & 67.8 & 72.3 & 84.9 & 94.8 & 52.9 & 78.7 \\
		P2Net~\cite{guo2019beyond} (ICCV'19) & 73.6 & 78.3 & 85.6 & 95.2 & - & - \\
		SCAL~\cite{chen2019self} (ICCV'19) & 72.3 & 74.8 & 89.3 & 95.8 & - & - \\
		DSA~\cite{zhang2019densely} (CVPR'19) & 75.2 & 78.9 & 87.6 & 95.7 & - & - \\
		DGNet~\cite{zheng2019joint} (CVPR'19) & - & - & 86.0 & 94.8 & 52.3 & 77.2 \\
		GCP~\cite{park2020relation} (AAAI'20) & 75.6 & 77.9 & 88.9 & 95.2 & - & - \\
		SAN~\cite{jin2020semantics} (AAAI'20) & 76.4 & 80.1 & 88.0 & \underline{96.1} & 55.7 & 79.2  \\
		ISP~\cite{zhu2020identity} (ECCV'20) & 74.1 & 76.5 & 88.6 & 95.3 & - & - \\
		GASM~\cite{he2020guided} (ECCV'20) & - & - & 84.7 & 95.3 & 52.5 & 79.5 \\
		RGA-SC~\cite{zhang2020relation} (CVPR'20) & \underline{77.4} & \underline{81.1} & 88.4 & \underline{96.1} & - & - \\
		HOReID~\cite{wang2020high} (CVPR'20) & - & - & 84.9 & 94.2 & - & - \\
		AMD~\cite{chen2021explainable} (ICCV'21) & - & - & 87.1 & 94.8 & - & - \\
		PGFL-KD~\cite{zheng2021pose} (ICCV'21) & - & - & 87.2 & 95.3 & - & - \\
		TransReID~\cite{he2021transreid} (ICCV'21) & - & - & \underline{89.5} & 95.2 & \textbf{67.4} & \textbf{85.3} \\
		PAT~\cite{li2021diverse} (CVPR'21) & - & - & 88.0 & 95.4 & - & - \\
		\hline
		MGN+R50 (UP-ReID) & \textbf{85.3} & \textbf{87.6} & \textbf{91.1} & \textbf{97.1} & \underline{63.3} & \underline{84.3} \\
		\shline
	\end{tabular}\\
	\label{tab:sota2}
\end{table*}

\section{Discussion about Hard Mining Strategy}
In Section 3.4 of the main body, we introduce our hard mining strategy in detail and experimentally prove its effectiveness in Section 4.5. Here we further discuss two points and give more insights about this design. The first one is that we choose hard positive samples and hard negative queues in a fixed way, which is an offline scheme instead of an online scheme. Would an online scheme be better? The second one comes from the positive samples selection. In Section 3.3 of the main body, we emphasize that all 2$M$ patch-level instances are partitioned from the input image $x$ actually. So, for each patch feature $q_i\in \mathcal{X}_{q}$, any of patch feature $k^+_p\in \mathcal{X}_{k}$ ($i,p \in \{1,...,M\}$) could be its positive sample. So, why do we have to choose patches at the same horizontal position instead of other patches as the positive samples?

\begin{figure}[h!]
	\centering
	\includegraphics[width=8.2cm, height=5.2cm]{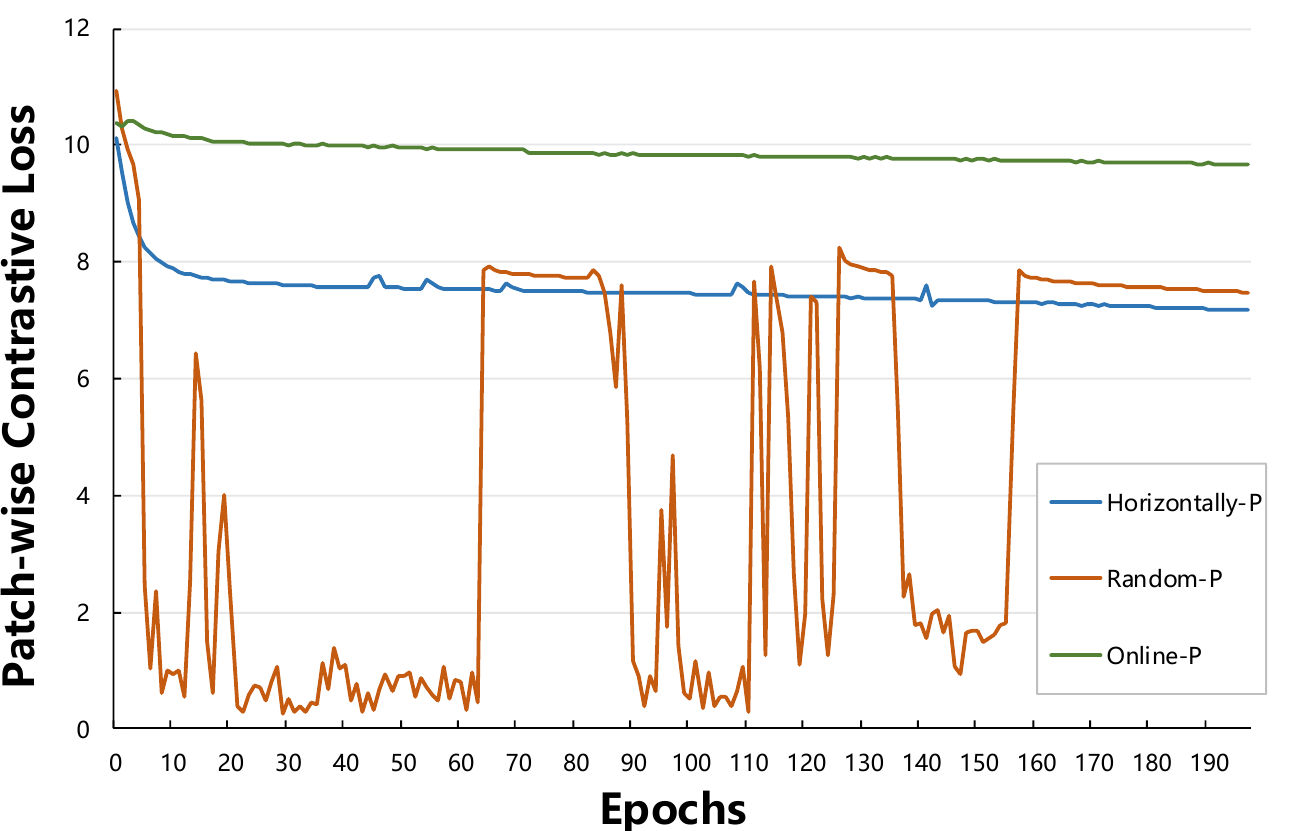}
	\setlength{\abovecaptionskip}{7pt}
	\vspace{-3mm}
	\caption{The curves of the patch-wise contrastive loss in different selection strategies. ``Horizontally-P'', ``Random-P'', and ``Online-P'' mean Horizontally Symmetric Positive Selection, Random Positive Selection, and Online Positive Selection, respectively.}
	\label{fig: patch-wise contrastive loss}
\end{figure}

To answer the aforementioned questions and verify the reasonableness of our selection strategy, we compare it with several other schemes. \textbf{\emph{Random Positive Selection}}: for patch $i$, we randomly select a patch partitioned from the same pedestrian but located differently as the positive sample. \textbf{\emph{Online Positive Selection}}: instead of finding a hard positive patch sample for each query patch $i$, we only select the hardest positive pair among all the $M\times M$ positive pairs. \textbf{\emph{Horizontally Symmetric Positive Selection}}: the proposed selection strategy wherein two horizontally symmetric patches are selected as a positive pair. Note that all three schemes have the same rule to select negative samples. We show the curves of the patch-wise contrastive loss in the intrinsic contrastive constraint under these three selection strategies in Figure~\ref{fig: patch-wise contrastive loss}.
As we can see, the loss value in the scheme of ``Random-P'' is unstable and cannot reach a convergence. On the other hand, the loss value in the scheme of ``Online-P'' converges extremely slowly. 

We analyze that the scheme of ``Random Positive Selection'' and ``Online Positive Selection'' suffer from misalignment and can not guarantee that the selected positive pairs have similar visual information. Take the ``Random Positive Selection'' as an example, for $q_i\in \mathcal{X}_{q}$, we randomly select $k^+_p\in \mathcal{X}_{k}$ as the corresponding positive sample. However, without any constraint, the visual information contained in $q_i$ and $k^+_p$ may be very different~(\eg, $q_i$ represents the head of a person, while $k^+_p$ represents the shoes), which has a negative impact on the pre-training process.

Our hard mining strategy~(\ie, Horizontally Positive Selection) is based on the prior knowledge that persons are horizontally symmetric, which assures that the positive pairs are semantically matched. This avoids the negative impact caused by misalignment on the pre-training process.

\section{Feature Visualization}
\begin{figure}[h!]
	\centering
	\includegraphics[width=8.2cm, height=6.2cm]{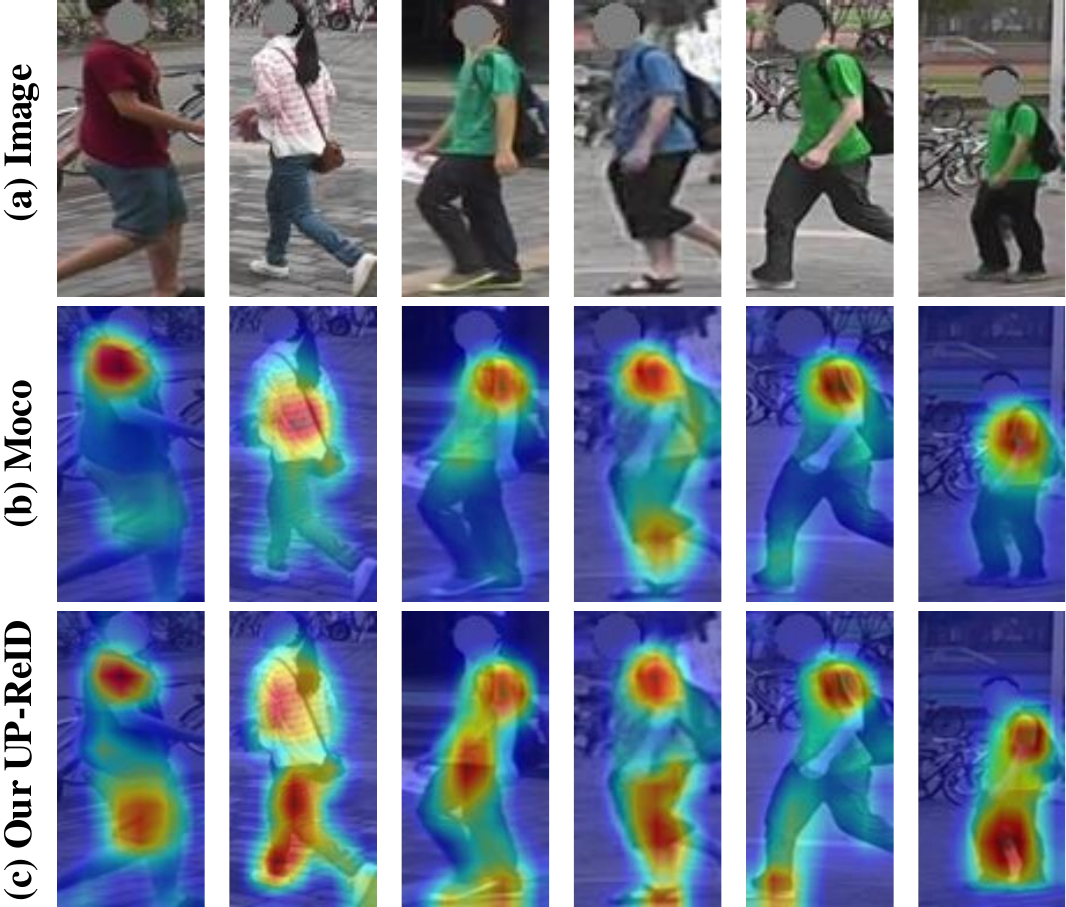}
	\setlength{\abovecaptionskip}{7pt}
	\vspace{-3mm}
	\caption{Visualization of the features corresponding to the Moco and our UP-ReID schemes.}
	\label{fig: Vis}
\end{figure}

As discussed in the main body, model pre-trained by our UP-ReID has better discriminative feature learning ability than that pre-trained by Moco. We fine-tune these two models in BOT~\cite{luo2019bag} on Market1501 for a few epochs, respectively. Then, we visualize the feature responses of our UP-ReID and Moco in Figure~\ref{fig: Vis}. As we can see, in the downstream tasks, UP-ReID pre-trained model could capture identity-related attributes (\eg, trouser color) and fine-grained features (\eg, shoes color) better than Moco pre-trained model, which demonstrates the effectiveness of the proposed designs, like the intrinsic contrastive constraint, in our UP-ReID.

\section{Broader Impacts}
As for positive impact, we demonstrate that a suitable pre-trained model can benefit downstream person ReID tasks with higher accuracy and faster convergence speed. This will improve efficiency and effectiveness of a series of ReID tasks and save human costs in these areas.

As for negative impact, many public ReID datasets are coming from unauthorized surveillance data, which may cause an invasion of privacy and other security issues. Thus, the collection process should be public and make sure that human subjects in the datasets are aware that they are being recorded. Strict regulation should also be established for ReID datasets to avoid ethical issues.

\end{document}